\def\year{2019}\relax
\documentclass[letterpaper]{article} 
\usepackage{aaai19}  
\usepackage{times}  
\usepackage{helvet}  
\usepackage{courier}  
\usepackage{url}  
\usepackage{graphicx}  
\usepackage{amsmath}
\usepackage{times}
\usepackage{xcolor}
\usepackage{soul}
\usepackage[utf8]{inputenc}
\usepackage[small]{caption}
\usepackage{booktabs}  
\usepackage{amsfonts}
\usepackage{bm}
\usepackage{graphicx}
\usepackage{tabu}
\usepackage{mathrsfs}
\usepackage{array}
\usepackage{booktabs}
\usepackage{multirow}
\usepackage{subfigure}
\usepackage{caption}
\DeclareMathOperator*{\argmax}{arg\,max}
\begin{document}
%
\title{Learning to Compose over Tree Structures via POS Tags}
\author{Gehui Shen \quad Zhi-Hong Deng\thanks{Corresponding author} \quad Ting Huang \quad Xi Chen\\
Key Laboratory of Machine Perception (Ministry of Education),\\
School of Electronics Engineering and Computer Science, Peking University,\\
Beijing 100871, China\\
{\tt \{jueliangguke, zhdeng, ht1221, mrcx\}@pku.edu.cn}}
\maketitle
\begin{abstract}
Recursive Neural Network (RecNN), a type of models which compose words or phrases recursively over syntactic tree structures, has been proven to have superior ability to obtain sentence representation for a variety of NLP tasks. However, RecNN is born with a thorny problem that a shared compositional function for each node of trees can't capture the complex semantic compositionality so that the expressive power of model is limited. In this paper, in order to address this problem, we propose Tag-Guided HyperRecNN/TreeLSTM (TG-HRecNN/TreeLSTM), which introduces hypernetwork into RecNNs to take as inputs Part-of-Speech (POS) tags of word/phrase and generate the semantic composition parameters dynamically. Experimental results on five datasets for two typical NLP tasks show proposed models both obtain significant improvement compared with RecNN and TreeLSTM consistently. Our TG-HTreeLSTM outperforms all existing RecNN-based models and achieves or is competitive with state-of-the-art on four sentence classification benchmarks. The effectiveness of our models is also demonstrated by qualitative analysis.
\end{abstract}

\section{Introduction}
Recently, as deep neural models are popular in NLP research community, learning distributed sentence representation becomes a basic but crucial problem for a variety of NLP tasks, including but not limited to sentence classification~\cite{kim14,tai15}, question qnswering~\cite{tan16,sgh}, and natural langauge inference~\cite{yu17,lin17}. 
In a common perspective, sentences are considered as sequences of words and recurrent neural networks (RNN) with long short-term memory (LSTM)~\cite{lstm} and convolutional neural networks (CNN)~\cite{kim14} are adapted to model sentences sequentially. However, this type of models can't always achieve the best performance because they ignore the syntactic structure. 

In contrast, RecNN~\cite{socher11} assigns a vector representation to each word at the leaf nodes of the pre-obtained syntactic parse tree structures and composes word/phrase pairs to get phrase representation at each non-leaf nodes recursively over trees. The final representation of root node is regarded as sentence representation. It is from a completely different perspective that natural language allows speakers to determine the meaning of a longer expression based on the meanings of its words and the rules used to combine them~\cite{socher12}. 
\begin{figure}
\centering
\includegraphics[height=0.16\textheight]{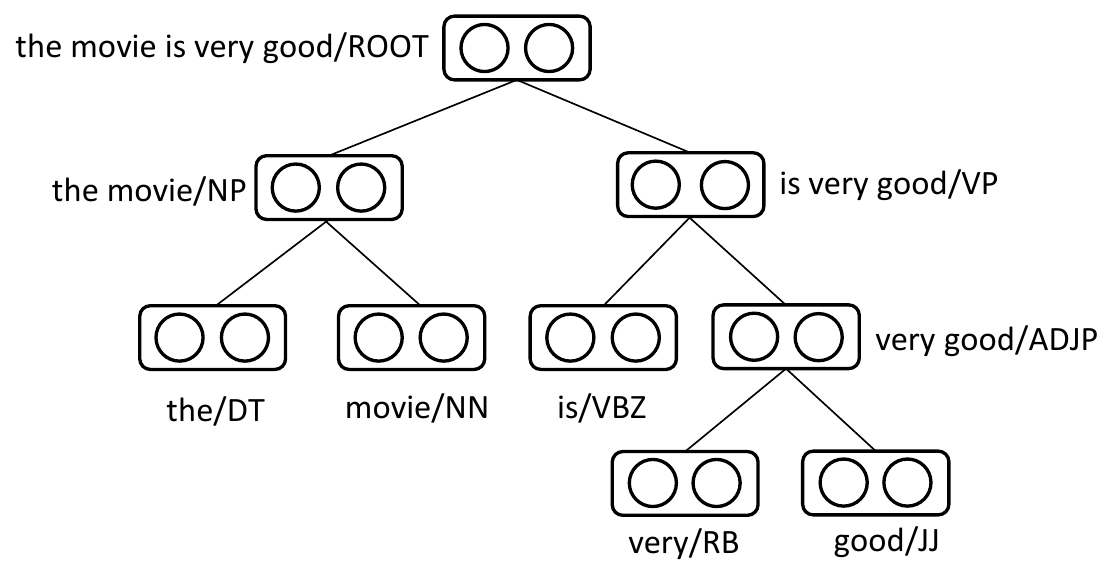}
\caption{An example of constituency parser tree. The words/phrases (e.g. movie, very good) and corresponding POS tags (e.g. NN, ADJP) of each node are illustrated.}
\end{figure}

Shortly after the standard RecNN is presented, researchers are aware that different from RNN, a group of shared parameters of semantic composition function limits the capacity of RecNN because different types of word/phrase pairs require pretty different composition rules. For example, the compositionality function of a \textit{verb-noun} pair should distinguish from the function of an \textit{adverb-adjective} pair intuitively. Therefore, some RecNN-based models are proposed for ~\cite{socher12,socher13a,socher13b,dong14,qian15,huang17} modeling the diversity and complexity of semantic composition. 
Overall speaking, these RecNN variants can be divided into two classes which partition composition function according to implicit and explicit rules respectively. The first class of models~\cite{socher12,socher13b,dong14} do not take syntactic information into consideration. Therefore, models have to determine a suitable composition function without any guidance, which makes the learning of models more difficult. The other class of models~\cite{socher13a,qian15,huang17} define an untied composition function for different Part-Of-Speech (POS) tags which represents syntactic roles of words and phrases. These models can select corresponding function according to tag information at each node. However the number of parameters are one or two orders of magnitude much larger than standard RecNN so that models easily suffer from overfitting.

In this paper, we propose \textbf{T}ag-\textbf{G}uided \textbf{H}yper RecNN/TreeLSTM (TG-HRecNN/TreeLSTM) to learn a dynamic semantic compositon function over tree structures with the help of POS tags. We introduce \textit{hypernetwork} framework~\cite{ha17} into RecNN/TreeLSTM. In the proposed models, a main RecNN/TreeLSTM whose parameters are tag-specific, is used to compose word/phrase pairs and learn sentence representation as ordinary RecNNs do. The purpose of hyper RecNN/TreeLSTM is to predict parameters of the main RecNN/TreeLSTM dynamically under the guidance of POS tags. Our work is inspired by recent progress in dynamic parameter prediction~\cite{bertinetto16,jia16,ha17} and the motivation is two-fold. \textbf{First}, to obtain tag-specific composition functions, without increasing the number of parameters significantly, a hypernetwork which takes the factors that determine the semantic composition rules as inputs and has a similar architecture with main RecNN, is a good choice. The factors include POS tags of nodes and hidden vectors which have been proved to be useful in previous works. \citeauthor{lpf17a}~\shortcite{lpf17a} conduct a pioneer work which firstly combines dynamic parameter prediction with RecNN. But their prediction network namely meta network, takes only hidden vector as inputs so that it is unable to capture syntactic information 
as discussed above. The empirical results show this work does not perform strongly enough so that it is meaningful to discovery a way to combine dynamic parameter prediction with RecNN as well as achieve excellent performance.

%
\textbf{Second}, we have observed the tag distribution is very imbalanced over datasets. For example, in Stanford Sentiment Treebank~\cite{socher13b} corpus, the frequencies of different POS tags differ from less than ten to tens of thousands. There are over 70 different types of tags in amount and if considering the combination of parent and child nodes, the tag configurations of nodes are much more. It is unrealistic to establish a composition function for each configuration explicitly. In case of overfitting due to discrete representation of POS tags, we use low-dimension distributed vectors to represent tags which we term tag embedding.

We focus on learning sentence representation over constituency parser tree in which each non-leaf node just has two children. We give an example for constituency tree in Figure 1. It should be noted when parsing sentences into constituency trees, POS tags of word/phrase are obtained simultaneously so no extra preprocessing is need for our models except parsing. In summary, our contributions are as follows:

\begin{enumerate}
\item To improve the semantic composition process over tree structures, we introduce \textit{hypernetwork} framework into RecNN and propose two novel models: TG-HRecNN and TG-TreeLSTM. Semantic composition parameters of our models are predicted dynamically under the guidance of POS tag information.
\item We design the \textit{information fusion layer} to incorporate POS tag and semantic information at each node to guide the composition process. The obvious gap between our models and DC-RecNN/TreeLSTM~\cite{lpf17a} manifests the effectiveness of tag information in dynamic parameter prediction. 
\item Experiments on five datasets for two typical NLP tasks show proposed models both obtain significant improvement compared with RecNN and TreeLSTM consistently. Specially, TG-HTreeLSTM outperforms all existing RecNN-based models and achieves or is competitive with state-of-the-art performance on four sentence classification benchmarks. The qualitative analysis illustrates how our model works.  

\end{enumerate}

\section{Related Work}

\subsection{Recursive Neural Networks}
Since RecNN~\cite{socher11} was proposed, several works focus on improving composition function over tree structures. \citeauthor{socher12}~\shortcite{socher12} replace vectors with matrix-vector pairs to represent nodes and matrix-vector multiplications are expected to model semantic composition flexibly and adaptively. \citeauthor{socher13b}~\shortcite{socher13b} utilize more complex bilinear neural tensor layer as composition function. \citeauthor{dong14}~\shortcite{dong14} improve RecNN by learning multiple composition functions whose outputs are summed up weightedly with self-adaptive weights. However, too much parameters make the learning of these models difficult. Besides these implicit ways to make composition function more flexible, \citeauthor{socher13a}~\shortcite{socher13a} and \citeauthor{qian15}~\shortcite{qian15} establish different composition function for each type of syntactic constituents. \citeauthor{qian15}~\shortcite{qian15} also propose a Tag Embedded RNN (TE-RNN) which firstly uses embedding vectors to represent POS tags of words/phrases and then takes tag vectors as inputs of composition function. A similar idea about exploiting POS tags has also been introduced into TreeLSTM~\cite{huang17}. \citeauthor{lpf17b}~\shortcite{lpf17b} and \citeauthor{lpf17c}~\shortcite{lpf17c} aim to address the non-compositional phenomenon and compose non-compositional phrases with a special function. 

In addition, there are other explorations on enhancing the RecNN. LSTM cell could also help to learn long-term dependencies over trees~\cite{tai15}. \citeauthor{teng17}~\shortcite{teng17} propose a bi-directional version of TreeLSTM.

\subsection{Dynamic Parameter Prediction}
The idea of dynamic parameter prediction that modifies the weights of one network by another is closely related to the concept of fast weights~\cite{js92} in which one network can produce context-dependent weight changes for a second network.  
Recently, this idea draws researchers's attention again because of the renaissance of deep neural networks.  \citeauthor{bertinetto16}~\shortcite{bertinetto16} attempt to learn example-dependent network weights for one-shot learning. \citeauthor{jia16}~\shortcite{jia16} introduce a framework which can dynamically generates CNN filters depended on network inputs. \citeauthor{ha17}~\shortcite{ha17} propose a hypernetwork framework to generate weights for recurrent networks. They can be seen as a form of relaxed weight-sharing in the time dimension. Due to the similarity between RNN and RecNN, we construct our model based on this framework. \citeauthor{lpf17a}~\shortcite{lpf17a} firstly employ the idea of dynamic parameter prediction to improve RecNN while the parameters are generated by exploiting limited information so state-of-the-art performance is not achieved. 

\section{Background: Hypernetwork}
Hypernetwork~\cite{ha17} framework is proposed to break the parameter sharing characteristic of recurrent networks. In this framework, there are two RNNs which are called hyper RNN and main RNN respectively. The former network is a standard RNN and is utilized to predict parameters of main RNN dynamically. Input sequences, such as sentences, are modeled by the latter as an ordinary RNN does. The update formulation of main RNN is given by:
\begin{equation} 
{\bf h}_t=\phi({\bf W}({\bf z}_w){\bf x}_t+{\bf U}({\bf z}_u){\bf h}_{t-1}+{\bf b}({\bf z}_b))
\end{equation}
All ${\bf z}_{(.)}$ vectors are outputs of hyper RNN and parameter matrices ${\bf W}$, ${\bf U}$ and bias ${\bf b}$ are functions of corresponding ${\bf z}_{(.)}$ where (.) $\in$ $\{w,u,b\}$. The input ${\bf x}_t \in \mathbb{R}^d$, hidden state ${\bf h}_t \in \mathbb{R}^h$ and nonlinearity $\phi$ are same as that in the standard RNN. The  update formulation of hyper RNN is similar with that of standard RNN:
\begin{equation} 
{\bf \widehat{h}}_t=\phi({\bf \widehat{W}}{\bf \widehat{x}}_t+{\bf \widehat{U}}{\bf \widehat{h}}_{t-1}+{\bf \widehat{b}})
\end{equation} 
\begin{equation}
{\bf \widehat{x}}_t = [{\bf x}_t; {\bf h}_{t-1}]
\end{equation} 
where ${\bf z}_{(.)} \in \mathbb{R}^z$ are linear functions of ${\bf \widehat{h}}_t$:
\begin{equation}
{\bf z}_{(.)} = {\bf W}_{\widehat{h}(.)}{\bf \widehat{h}}_t + {\bf b}_{\widehat{h}(.)}
\end{equation}
 If directly projecting ${\bf z}_{x}$ into the matrix ${\bf W}$, it may be not practical because we have to maintain a $h \times d \times z$ learnable tensor so that the memory usage becomes too large for real problems. An approximate mechanism is to revise equation 1 in the following way:
\begin{equation}
{\bf h}_t=\phi({\bf z}_w \odot {\bf W}{\bf x}_t+{\bf z}_u\odot{\bf U}{\bf h}_{t-1}+{\bf z}_b)
\end{equation}
We can see that now ${\bf z}_{(.)}$ modify the corresponding parameters by scaling each row of weight matrix linearly by an element in vector. Although in this way the degrees of freedom of parameter prediction process is reduced, the memory usage becomes available. It should be noted that in practice the dimension of hypernetwork is much lower than that of main network which means the total number of model parameters will not increase significantly.  

\begin{figure}
\centering
\includegraphics[height=0.27\textheight]{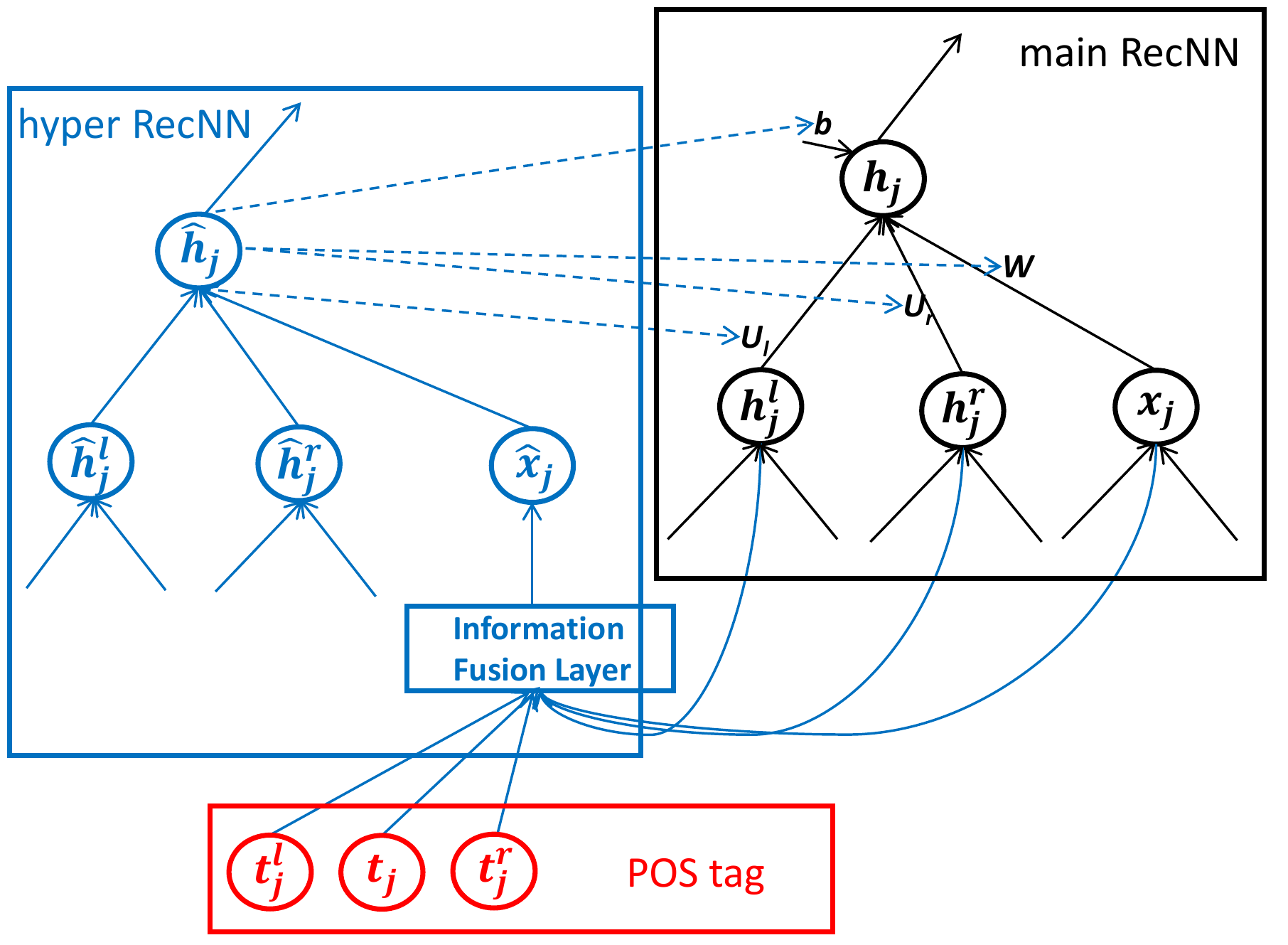}
\caption{An overview of TG-HRecNN at node $j$. The main RecNN and hyper RecNN are painted in black and blue respectively. The dotted arrows represent dynamic parameter prediction process. The POS tags are inputs of information fusion layer of hyper RecNN.}
\end{figure}

\section{Proposed Models}
RecNNs can model sentences over tree structures. They use a group of shared parameters to compose word/phrase pairs recursively which limits the expressive power of models. Inspired by the fact that word/phrase pairs with different POS tags require different semantic composition functions, we introduce the hypernetwork framework into vanilla RecNN and TreeLSTM and propose \textit{Tag-Guided Hyper RecNN} and \textit{Tag-Guided Hyper TreeLSTM} respectively. Both of them consist of a hyper network and a main network as hyper RNN. Figure 2 shows the illustration of proposed model. At each nodes of the tree, adaptive composition parameters are predicted dynamically according to POS tag information. 


\subsection{Tag-Guided HyperRecNN (TG-HRecNN)}
For a non-leaf node $j$ in constituency parse trees, the RecNN obtains its hidden state ${\bf h}_j \in \mathbb{R}^h$ by composing the hidden states of its left child and right child, namely ${\bf h}_j^l$ and ${\bf h}_j^r$ respectively. The composition function is a simple affine transformation as follows:
\begin{equation}
{\bf h}_j=f\left(\bf U\begin{bmatrix}{\bf h}_j^l\\{\bf h}_j^r\end{bmatrix}+b\right)
\end{equation}
where $\bf U$ $\in \mathbb{R}^{h\times 2h}$ is the parameter matrix and $\bf b$ $\in \mathbb{R}^h$ is the bias. $f = tanh$ is a standard element-wise nonlinearity. 

TG-HRecNN consists of a main RecNN and a hyper RecNN. The hyper RecNN is similar to a standard RecNN whose composition function is given by: 
\begin{equation}
{\bf \widehat{h}}_j=\phi({\bf \widehat{W}}{\bf \widehat{x}}_j+{\bf \widehat{U}}\begin{bmatrix}{\bf \widehat{h}}_j^l\\{\bf \widehat{h}}_j^r\end{bmatrix}+{\bf \widehat{b}})
\end{equation}
Compared to vanilla RecNN, there is an additional input ${\bf \widehat{x}}_j \in \mathbb{R}^{\widehat{d}}$ which contains POS tag information to determine the semantic composition at current node. We design an \textit{information fusion layer} to compute ${\bf \widehat{x}}_j$ and we delay introducing this layer until Sec 4.3. ${\bf \widehat{h}}_j \in \mathbb{R}^{\widehat{h}}$ denotes the hidden state and ${\bf \widehat{W}} \in \mathbb{R}^{\widehat{h}\times \widehat{d}}$, ${\bf \widehat{U}} \in \mathbb{R}^{\widehat{h}\times 2\widehat{h}}$ and ${\bf \widehat{b}} \in \mathbb{R}^{\widehat{h}}$ are learnable parameters.

Similar to the main RNN in equation 5, the compostition function of main RecNN has dynamic parameters:

\begin{equation}
{\bf h}_j=f\left({\bf z}_u \odot {\bf U}\begin{bmatrix}{\bf h}_j^l\\{\bf h}_j^r\end{bmatrix}+{\bf z}_b\right)
\end{equation}
All ${\bf z}_{(.)} \in \mathbb{R}^z$ are computed through linear transformation as equation 4 where (.) $\in$ $\{w,b\}$:
\begin{equation}
{\bf z}_{(.)} = {\bf W}_{\widehat{h}(.)}{\bf \widehat{h}}_j + {\bf b}_{\widehat{h}(.)}
\end{equation}

\subsection{Tag-Guided HyperTreeLSTM (TG-HTreeLSTM)}
\citeauthor{tai15}~\shortcite{tai15} adopts a standard LSTM cell to TreeLSTM cell which can be used as composition function over constituency parse trees. The hidden state ${\bf h}_j$ now is calculated as follows:
\begin{small}
\begin{equation}
\begin{bmatrix} {\bf g}_j \\ {\bf i}_j \\ {\bf f}_j^l \\ {\bf f}_j^r \\{\bf o}_j \end{bmatrix} =  \begin{bmatrix} tanh \\ \sigma \\ \sigma \\ \sigma \\ 
\sigma\end{bmatrix}\left({\bf W} {\bf x}_j + {\bf U}\begin{bmatrix} {\bf h}_j^l \\ {\bf h}_j^r\end{bmatrix} + {\bf b}\right)
\end{equation}
\end{small}
\begin{equation}
{\bf c}_j = {\bf i}_j \odot {\bf g}_j + {\bf f}_j^l \odot {\bf c}_j^l + {\bf f}_j^r \odot {\bf c}_j^r
\end{equation}
\begin{equation}
{\bf h}_j =  {\bf o}_j \odot tanh({\bf c}_j)
\end{equation}
where ${\bf c}_j \in \mathbb{R}^h$ denotes the memory cell. 
${\bf x}_j \in \mathbb{R}^d$ is the input of node $j$ which is word embedding at leaf nodes or zero at non-leaf nodes. $\sigma$ denotes sigmoid function 
and $\odot$ denotes elementwise multiplication. ${\bf i}_j, {\bf f}_j, {\bf o}_j$ are termed \textit{input gate}, \textit{forget gate} and \textit{output gate} respectively. The superscript $l$ and $r$ represent the left child and right child respectively. For binarized trees this model computes two untied forget gates for each children. All matrices ${\bf W} \in \mathbb{R}^{5h \times d}$, ${\bf U}^l \in \mathbb{R}^{5h \times h}$,  ${\bf U}^r \in \mathbb{R}^{5h \times h}$ and bias ${\bf b} \in \mathbb{R}^{5h}$ are learnable parameters. 

TG-HTreeLSTM also consists of a main TreeLSTM and a hyper TreeLSTM. The composition formulation of hyper TreeLSTM is almost identical to TreeLSTM except the definition of ${\bf \widehat{x}} \in \mathbb{R}^{\widehat{d}}$ which will be described in Sec 4.3:
\begin{small}
\begin{equation}
\begin{bmatrix} {\bf \widehat{g}}_j \\ {\bf \widehat{i}}_j \\ {\bf \widehat{f}}_j^l \\ {\bf \widehat{f}}_j^r \\{\bf \widehat{o}}_j \end{bmatrix} =  \begin{bmatrix} tanh \\ \sigma \\ \sigma \\ \sigma \\ 
\sigma\end{bmatrix}\left({\bf \widehat{W}} {\bf \widehat{x}}_j + {\bf \widehat{U}}\begin{bmatrix} {\bf \widehat{h}}_j^l \\ {\bf \widehat{h}}_j^r\end{bmatrix} + {\bf \widehat{b}}\right)
\end{equation}
\end{small}
\begin{equation}
{\bf \widehat{c}}_j = {\bf \widehat{i}}_j \odot {\bf \widehat{g}}_j + {\bf \widehat{f}}_j^l \odot {\bf \widehat{c}}_j^l + {\bf \widehat{f}}_j^r \odot {\bf \widehat{c}}_j^r
\end{equation}
\begin{equation}
{\bf \widehat{h}}_j =  {\bf \widehat{o}}_j \odot tanh({\bf \widehat{c}}_j)
\end{equation}
where ${\bf \widehat{W}} \in \mathbb{R}^{5\widehat{h}\times \widehat{d}}$, ${\bf \widehat{U}} \in \mathbb{R}^{5\widehat{h}\times 2\widehat{h}}$ and ${\bf \widehat{b}} \in \mathbb{R}^{5\widehat{h}}$ are learnbale parameters. Then we use ${\bf \widehat{h}}_j \in \mathbb{R}^{\widehat{h}}$ to dynamically predict parameters of main LSTM:
\begin{small}
\begin{equation}
\begin{bmatrix} {\bf g}_j \\ {\bf i}_j \\ {\bf f}_j^l \\ {\bf f}_j^r \\{\bf o}_j \end{bmatrix} =  \begin{bmatrix} tanh \\ \sigma \\ \sigma \\ \sigma \\ 
\sigma\end{bmatrix}\left({\bf z}_w \odot {\bf W} {\bf x}_j + {\bf z}_u \odot {\bf U}\begin{bmatrix} {\bf h}_j^l \\ {\bf h}_j^r\end{bmatrix} + {\bf z}_b \right)
\end{equation}
\end{small}
All ${\bf z}_{(.)} \in \mathbb{R}^z$ are computed through linear transformation by equation 9 where (.) $\in$ $\{w,u,b\}$ and hidden state ${\bf h}_j$
can be obtained by equation 11-12.

\subsection{Information Fusion Layer}
To enable the hyper network to guide the semantic composition of the main network, we incorporate the syntactic information and semantic representation into ${\bf \widehat{x}}$, the input of hyper network at each node in equation 6 and 12.  For a non-leaf node $j$, we refer to ${\bf t}_j \in \mathbb{R}^t$ as its tag embedding and the tag embeddings of its children are denoted by ${\bf t}_j^l, {\bf t}_j^r$ respectively. We consider ${\bf t}_j, {\bf t}_j^l, {\bf t}_j^r$ as syntactic information which determine the composition function at each node. We make use of the semantic information about the node as well. In equation 3, the hyper RNN~\cite{ha17} utilizes the hidden state of last time step ${\bf x}_t$ and current input ${\bf h}_{t-1}$ of main RNN. However, the input ${\bf x}_j$ in non-leaf node $j$ is zero. To fill this gap, we resort to the head-lexicalized~\cite{lexical} in PCFG parser. A non-leaf node is associated with a head word which is the head word of one of its children according to pre-defined rules. \citeauthor{teng17}~\shortcite{teng17} firstly exploit it in neural network in a soft gated way. We calculate the head word ${\bf x}_j$ for a non-leaf node $j$ in a similar way while utilizing tag embeddings as additional inputs:
\begin{equation}
{\bf a}_j = \sigma({\bf W}_{head}[{\bf t}_j;{\bf t}_j^l;{\bf t}_j^r;{\bf x}_j^l;{\bf x}_j^r]+{\bf b}_{head})
\end{equation}
\begin{equation}
{\bf x}_j = {\bf a}_j \odot {\bf x}_j^l + (1-{\bf a}_j) \odot {\bf x}_j^r
\end{equation}
where ${\bf x}_j^l, {\bf x}_j^r \in \mathbb{R}^d$ are the head word of two children nodes. ${\bf a}_j$ controls the composition of head words adaptively. ${\bf W}_{head} \in \mathbb{R}^{d \times (3t+2d)}$ and ${\bf b}_{head} \in \mathbb{R}^{d}$ are learnable parameters. Then we can calculate the ${\bf \widehat{x}}_j$ with two heuristic strategies. The first is to directly concatenate all tag embeddings and semantic representations as follows:
\begin{equation}
{\bf \widehat{x}}_j = ReLU({\bf W}_{\widehat{x}}[{\bf t}_j;{\bf t}_j^l;{\bf t}_j^r;{\bf x}_j;{\bf h}_j^l;{\bf h}_j^r]+{\bf b}_{\widehat{x}})
\end{equation}
where $ReLU$ is Rectified Linear Units as nonlinearity. ${\bf W}_{\widehat{x}} \in \mathbb{R}^{\widehat{d} \times (3t+2h+d)}$ and ${\bf b}_{\widehat{x}} \in \mathbb{R}^{\widehat{d}}$ are learnable parameters.
The other strategy is to project tag embeddings and semantic representations separately and then operate an element-wise multiplication between them:
\begin{equation}
{\bf \overline{t}}_j = [{\bf t}_j;{\bf t}_j^l;{\bf t}_j^r], \indent {\bf \overline{s}}_j = [{\bf x}_j;{\bf h}_j^l;{\bf h}_j^r] 
\end{equation}
\begin{equation} 
{\bf \widehat{x}}_j = ReLU({\bf W}_{\widehat{x}}^{t} {\bf \overline{t}}_j \odot{\bf W}_{\widehat{x}}^{s}{\bf \overline{s}}_j +{\bf b}_{\widehat{x}})
\end{equation}
where ${\bf W}_{\widehat{x}}^t \in \mathbb{R}^{\widehat{d} \times 3t}, {\bf W}_{\widehat{x}}^s \in \mathbb{R}^{\widehat{d} \times (2h+d)} $ and ${\bf b}_{\widehat{x}} \in \mathbb{R}^{\widehat{d}}$ are learnable parameters. We term these two strategies \textit{concat} and \textit{multi} for short respectively.

\begin{table}
\small
\begin{center}
\begin{tabular}{cccccccc}

\toprule[1pt]
\bf Dataset& $|\mathcal{N}|$ & \textbf{Test} &\textbf{Class} &$L_{avg}$&$|\mathcal{V}|$&$|\mathcal{T}|$\\ \hline \specialrule{0em}{1.3pt}{1.3pt}
SST& 11855& 2210& 5/2& 18 &21K& 71\\
MR& 10662& CV& 2&22&19K&72\\
SUBJ& 10000& CV& 2&21&21K&71\\
TREC& 5952&500& 6& 10&10K&66\\
SICK& 9927& 4927&3&10 &2K&40\\
\toprule[1pt]
\end{tabular}
\end{center}
\caption{Statistics of five datasets used in this paper. $|\mathcal{N}|$ denotes the size of dataset. \textbf{Test} denotes the size of testset where CV means there was no standard train/test split and thus 10-fold CV was used. \textbf{Class} denotes the number of categories. $L_{avg}$ denotes the average length of sentences. $|\mathcal{V}|$ denotes the size of vocabulary. $|\mathcal{T}|$ denotes  the number of different word/phrase tags in datasets.}
\end{table}

\begin{table*}[htb]
\small
\begin{center}
\begin{tabular}{lcccccc} 
\toprule[1pt]
\bf Model & SST-1 & SST-2 & MR & SUBJ & TREC & SICK \\ \hline \specialrule{0em}{1.3pt}{1.3pt}
CNN~\cite{kim14} &48.0&88.1&81.5&93.4&93.6&- \\ 
Bidirectional LSTM~\cite{tai15} &49.1&87.5&-&-&-&- \\ \hline \specialrule{0em}{1.3pt}{1.3pt}
RecNN~\cite{socher11} &43.2&82.4&76.4&91.8&90.2& 74.9\\
MV-RNN~\cite{socher12} &44.4&82.9&-&-&-&75.5\\
RNTN~\cite{socher13b} &45.7&86.4&-&-&-&76.9\\
AdaMC-RNN~\cite{dong14} &45.8&87.1&-&-&-&-\\
TG-RNN~\cite{qian15} &46.1&86.2&76.4&-&- &-\\
TE-RNN~\cite{qian15} &47.8&86.5&77.9&-&- &-\\
TreeLSTM~\cite{tai15} & 51.0& 88.0 & 81.2& 93.2&93.6&77.5\\  
AdaHT-LSTM~\cite{lpf17b} &50.2&87.8&81.9&94.1&-&-\\  
iTLSTM~\cite{lpf17c} &51.2&88.2&82.5&94.5&- &-\\ 
TE-LSTM~\cite{huang17} & 52.6 & 89.6 & 82.2& - & -&-\\
BiTreeLSTM~\cite{teng17} & 53.5 & 90.3 &-&-&94.8 &-\\ \hline \specialrule{0em}{1.3pt}{1.3pt}
DC-RecNN~\cite{lpf17a} &-&86.1&80.2&93.5&91.2 &77.9\\  
DC-TreeLSTM~\cite{lpf17a} &-&87.8&81.7&93.7&93.8 &80.2\\
\hline \specialrule{0em}{1.3pt}{1.3pt}
TG-HRecNN+concat (Proposed) &49.6&87.1&81.1&93.8&93.4&78.3\\
TG-HRecNN+multi (Proposed)&49.3&87.2&80.9&93.7&93.6&77.5\\
TG-HTreeLSTM+concat (Proposed) &\textbf{53.7}&90.2&\textbf{82.9}&94.7&95.0 &\textbf{83.6}\\
TG-HTreeLSTM+multi (Proposed)&53.2&\textbf{90.4}&82.6&\textbf{94.9}&\textbf{95.8}&83.3\\
\bottomrule[1pt] 
\end{tabular}
\end{center}
\caption{Accuracies of baseline models (in the first group), proposed models (in the last group) and previous RecNN-based models (in the second and third group) on five datasets. The \textit{concat} and \textit{multi} denotes two different strategies in information fusion layer described in Sec 4.3.
}
\end{table*}

\subsection{Output Layer}
Given a sentence with its parser tree, proposed models can compute hidden state of each node over the tree recursively. The hidden state ${\bf h}_{j}$ computed by proposed models can be regarded as the representation of the phrase spanned by node $j$. Specially, we use the hidden state of root node ${\bf h}_{root}$ as the sentence representation and apply it to two realistic NLP tasks. We utilize different output layers for two tasks.

For \textit{sentence classification}, we should predict a label $\hat{y}$ from a pre-defined class set $\mathcal{Y}$ 
for a sentence $x$. We directly feed the sentence representation ${\bf h}_{x}$ into a softmax classifier.
$$\hat{p}_{\theta}(y|x)=softmax({\bf W}_s{\bf h}_x+{\bf b}_s)$$

For \textit{text semantic matching}, we deal with a classification problem about sentence pairs. Given two sentences $s$ and $t$, we need to predict a label $\hat{y}$ which represents the relation between them. We firstly obtain their representations ${\bf h}_{s}$, ${\bf h}_{t}$ with a parameter shared TG-HRecNN/TreeLSTM and then combine the features in this way:
$${\bf h}_{x}=[{\bf h}_{s} \odot {\bf h}_{t};|{\bf h}_{s} - {\bf h}_{t}|]$$
We feed it into a network of one hidden layer  with ReLU activation before into softmax classifier:
$${\bf h}_{mlp}=ReLU({\bf W}_{mlp}{\bf h}_{x}+{\bf b}_{mlp})$$
$$\hat{p}_{\theta}(y|x)=softmax({\bf W}_{s}{\bf h}_{mlp}+{\bf b}_{s})$$
The training objective for two tasks is to minimize the cross-entropy of the predicted and true label distributions:
$$loss=-\frac{1}{|D|}\sum_{k=1}^{|D|}log \hat{p_{\theta}}(y^{(k)}|x^{(k)})$$
where $|D|$ is the number of training samples. $x^{(k)}$ and $y^{(k)}$ are the \textit{k}-th sample and label in dataset respectively. Then the prediction is given in this way:
$$\hat{y}=\argmax_{y} \hat{p}_{\theta}(y|x)$$

\section{Experiments}
\subsection{Datasets}
To evaluate the effectiveness of proposed models, we conduct experiments on four benchmarks for sentence classification and SICK dataset for text semantic matching:
\begin{itemize}
\item {\bf SST}: Stanford Sentiment Treebank~\cite{socher13b} for sentiment classification. SST-1 denotes the evaluation with fine-grained labels \textit{(very positive, positive, neutral, negative, very negative)} and SST-2 denotes the evaluation with binary labels by neglecting the \textit{neutral} samples during test. During training, we also utilize the phrase-level labels as previous works do.\footnote{http://nlp.stanford.edu/sentiment/}
\item {\bf MR}: Movie reviews with two polarity classes. \textit{(positive/negative)}~\cite{mr}\footnote{https://www.cs.cornell.edu/people/pabo/movie-review-data/}
\item {\bf SUBJ}: Subjectivity datasets with two classes. \textit{(subjective/objective)}~\cite{subj}\footnote{https://www.cs.cornell.edu/people/pabo/movie-review-data/}
\item {\bf TREC}: TREC question dataset with six question classes (e.g. \textit{location}).~\cite{trec}\footnote{http://cogcomp.org/Data/QA/QC/}
\item {\bf SICK}: Sentences Involving Compositional Knowledge for text entailment with three classes (\textit{entailment, contradiction, neutral}).~\cite{sick}\footnote{http://clic.cimec.unitn.it/composes/sick.html}
\end{itemize}
The detailed dataset statistics are listed in Table 1.
\subsection{Implementation Details}
In all experiments, we initialize word embeddings with 300-dimensional Glove 840B vectors\footnote{http://nlp.stanford.edu/projects/glove/} 
~\cite{glove}. We only fine-tune word embeddings on SST during training. We use AdaGrad~\cite{adagrad} optimizer with an initial learning rate of 0.05. The hidden size of main networks $h$ is 150. The hidden size of hyper networks $\hat{h}$ is 50 and the input size of hyper networks $\hat{d}$ is 100. The dimension of tag embedding is 50. We apply dropout on both embedding and output layer with a dropout rate of 0.5. Recurrent dropout~\cite{recdrt} with a dropout rate of 0.25 for main networks is applied for sentence classification and 3e-5 L2-regularization is applied for text semantic matching. The minibatch size is always 50. We obtain the constituency parser trees and POS tags of word/phrases using Stanford Parser~\cite{parser}. The code is implemented with Theano~\cite{theano}. 
\begin{table*}[htb]
\small
\begin{center}
\begin{tabular}{lccccc}
\toprule[1pt]
\bf Model & SST-1 & SST-2 & MR & SUBJ & TREC \\ \hline \specialrule{0em}{1.3pt}{1.3pt}
AdaSent~\cite{zhao15} &-&-&\textbf{83.1}&\textbf{95.5}&92.4\\
d-TBCNN~\cite{mou15} &51.4&87.9&-&-&96.0\\
DSCNN-Pretrain~\cite{zhang16} &50.6&88.7&82.2&93.9&95.6\\
BLSTM-2DCNN~\cite{zhou16} &52.4&89.5&82.3&94.0&\textbf{96.1}\\
NTI~\cite{yu17} &53.1&89.3&-&-&-\\
BCN+Char+CoVe$^\ast$~\cite{Socher17}&\textbf{53.7}&90.3&-&-&95.8\\
TG-HTreeLSTM (Proposed)&\textbf{53.7}&\textbf{90.4}&82.9&94.9&95.8\\
\bottomrule[1pt] 
\end{tabular}
\end{center}
\caption{Comparison between the proposed model and non RecNN-Based state-of-the-art models for sentence classification. The symbol $^\ast$ indicates besides GloVe this model also uses CoVe word embeddings which are trained with external resources.}
\end{table*}

\begin{table*}
\small
\begin{center}
\begin{tabular}{ccc} 
\toprule[1.0pt]
\textbf{Dimensions} & \textbf{Tag Types} & \textbf{Examples for Phrases to Compose} \\ \hline \specialrule{0em}{1.3pt}{1.3pt}
16-\textit{th} &JJ+NN& aching+beauty, perfect+film, recent+favourite, implausible+situation\\
30-\textit{th} & DT+NN & a+movie, this+examination, the+film, no+picture \\ 
62-\textit{nd} &NP+PP& the classic films+of Jean Renoir, the most powerful thing+in life, every opportunity+for a breakthrough\\ 
79-\textit{th} & ADJP+NN & most impossibly dry+account, far more thoughtful+film, weak or careless+performance\\ 
132-\textit{nd} &IN+NP& of+mystery and quietness, in+his bratty character, in+a mess of purposeless violence \\ 
\bottomrule[1.0pt] 
\end{tabular}
\end{center}
\caption{Some interpretable dimensions of ${\bf z}_u$ with the types of word/phrase pairs of the nodes where dynamic composition parameters have large scale in corresponding rows. The last column gives several examples for each type of pairs. ``+'' splits the tags or phrases of pairs.}
\end{table*}

\subsection{Sentence Classification}
We first compare proposed models with RecNN-based models as well as baseline models. Then we make comparison between TG-HTreeLSTM and non RecNN-based state-of-the-art models.

\subsubsection{Comparison with RecNN-Based Models} 
The experimental results about this comparison are displayed in columns 2 to 6 in Table 2. \textbf{Firstly}, we find that TG-HRecNN and TG-HTreeLSTM both outperform RecNN and TreeLSTM on all datasets. Specially, compared with RecNN and TreeLSTM, TG-HRecNN and TG-HTreeLSTM obtain about 5.4\%/4.8\% and 2.7\%/2.4\% improvements on SST-1/SST-2 respectively which are much greater than those on other datasets. We think it is beacuse there are phrase-level labels in SST so that the dynamic parameter prediction process can get more supervision during training. The vanilla RecNN are more easily boosted by dynamic parameter prediction and TG-HRecNN is competitive with strong baselines CNN and BiLSTM surprisingly. \textbf{Secondly}, compared with TG/TE-RNN and TE-LSTM which also exploit POS tags to enhance the expressive power of semantic composition function, our models are still superior to them. This means that to guide the semantic composition function dynamically with POS tags the hypernetwork framework is more effective. Proposed models are also much stronger than AdaMC-RNN, AdaHT-LSTM and iTLSTM which uses different composition function in a self-adaptive way although the last two models expliot external knowledge about idiom. \textbf{Thirdly}, compared with DC-RecNN/TreeLSTM, the results again demonstrate the superiority of our models. TG-HRecNN and TG-HTreeLSTM outperform DC-RecNN and DC-TreeLSTM with about 1\%-2.5\% accuracy on SST, MR and TREC respectively while on SUBJ our models also outperform them by a small margin. We think it can confirm our conjecture that POS tags are useful information for dynamic parameter prediction. Especially, DC-TreeLSTM has only similar performance with CNN without tag information. \textbf{Lastly}, TG-HTreeLSTM beats all RecNN-based models including bidirectional TreeLSTM (BiTreeLSTM)~\cite{teng17} which is powerful with the help of top-down information flow. Compared with BiTreeLSTM, the most competitive RecNN variant, TG-HTreeLSTM has much less parameters (0.84M versus 1.30M).

In addition, the two strategies \textit{concat} and \textit{multi} introduced for information fusion layer both can be more effective in some specific settings. 

\subsubsection{Comparison with non RecNN-Based State-of-the-art Models}
In table 3, we list the accuracies of our TG-HTreeLSTM model and state-of-the art models for sentence classification. Overall, TG-HTreeLSTM has consistently strong performances on four datasets and achieve the best scores on SST. We can find AdaSent~\cite{zhao15} performs very well on MR and SUBJ and keeps the state-of-the-art on the two datasets since 2015. Nevertheless, the gap between TG-HTreeLSTM and AdaSent is 0.2\% on MR and 0.6\% on SUBJ which are relatively small. On TREC, TG-HTreeLSTM is also competitive with the best BLSTM-2DCNN~\cite{zhou16} model with a 0.3\% gap. This comparison shows that TG-HTreeLSTM has excellent generalization ability. 

It should be emphasized that although BCN+Char+CoVe~\cite{Socher17} has almost the same results as TG-HTreeLSTM on SST, it uses extra CoVe word embeddings which are obtained by training a machine translation model. If combing with CoVe our model probably performs better.      

\subsection{Text Semantic Matching}
The last column in Table 2 summarizes the performance of different models on SICK. It should be noted that the aim of this experiment is to prove that proposed models can improve RecNN/TreeLSTM for different NLP tasks instead of pursuing state-of-the-art performance. So we only compare sentence encoding-based models which encode two sentences into vectors and then classify as described in Sec 4.4. TG-HRecNN and TG-HTreeLSTM achieve 3.4\%/6.1\% improvements than RecNN and TreeLSTM respectively. Although TG-HRecNN is only competitive with DC-RecNN, TG-HTreeLSTM shows effective performance in this task and outperforms DC-TreeLSTM with 3.4\% accuracy.

\subsection{Qualitative Analysis}
To explain the effectiveness of proposed models, we conduct experiments to examine how the hyper RecNN predicts the composition parameter of main RecNN dynamically. As we describe in Sec 4, hyper RecNN modifies the parameters by scaling each row of the parameter matrix. So we examine the value of each dimension of ${\bf z}_u$ in equation 8 which is output by hyper RecNN and determines the final composition parameter of main RecNN on the test set of SST. We find the occurrence of large value of some dimensions of ${\bf z}_u$ is dominated by nodes with specific tag types. Table 4 illustrates some examples of these interpretable dimensions which supports that proposed models can predict reasonable composition parameter for different types of word/phrase pairs by enlarging different rows of the parameter matrix. 

\begin{figure}[htb]
\centering
\includegraphics[height=0.25\textheight]{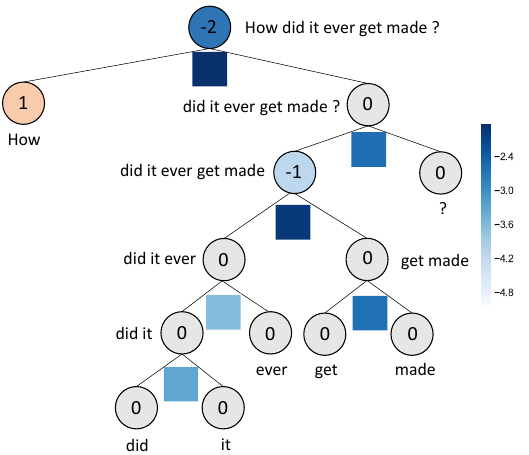}
\caption{The visualization of parser tree and values of the 13-\textit{rd} dimension of ${\bf z}_u$ for sentence ``How did it ever get made?''. -2, -1, 0, 1, 2 denote 5 sentiment labels (very negative to very positive) respectively in circles. The color in the squares below each non-leaf node represents values of ${\bf z}_u$. High intensity represent large values.} 
\end{figure}

We also find the occurrence of large value in some dimensions is dominated by nodes with specific sentiment polarity. In figure 3, we give a sample to visualize the behaviours of 13-\textit{rd} dimension which is sensitive to \textit{negative} sentiment. The label of the whole sentence is \textit{very negative} while labels of phrases are \textit{neutral} except ``did it ever get made''. The values of this dimension get much larger at two nodes with \textit{negative} label than those at other nodes. Although during test no label can be seen, this dimension of ${\bf z}_u$ entails sentiment information so that our model can give a correct prediction with adaptive composition parameters.

\section{Conclusion}
In this paper, we introduce hypernetwork framework into RecNNs to address the problem caused by shared composition parameter. We propose two novel RecNN variants in which a hyper RecNN taking as inputs POS tag information predicts the composition parameter of main RecNN dynamically. An information fusion layer is designed to incorporate POS tag and semantic information for parameter prediction. Our models beat all RecNN-based models on five datasets for sentence classification and text semantic matching. Proposed TG-HTreeLSTM achieves or is competitive with state-of-the-art on four sentence classification benchmarks. We also give qualitative analysis to explain why our models work well.

Experimental results show that proposed models are able to encode a sentence into powerful distributed representation, which will benefit many NLP tasks. In future work, we will employ our models as sentence encoders and apply encoded sentence embedding to high level tasks, such as document classification and reading comprehension. We will also explore the effectiveness of other syntax information for guiding dynamic parameter prediction. 
\bibliographystyle{aaai}
\bibliography{tg_recnn}
\end{document}